\documentclass[11pt]{article}

\usepackage[letterpaper, top=1.5in, bottom=1.2in, left=1.25in, right=1.25in]{geometry}
\usepackage{amsmath, amssymb, amsfonts}
\usepackage{graphicx}
\usepackage{microtype}
\usepackage{lmodern}
\usepackage{hyperref}
\usepackage{booktabs}
\usepackage{algorithm}
\usepackage{algpseudocode}
\usepackage{caption}
\usepackage[T1]{fontenc}
\usepackage{subcaption}
\usepackage{setspace}
\usepackage{enumitem}
\usepackage{xcolor}
\usepackage{tikz}
\usepackage{standalone}
\usepackage{fix-cm}
\usepackage{pgfplots}
\usepackage{bbm}

\usetikzlibrary{shapes.misc, shadows, arrows.meta, positioning, shapes.geometric, fit, backgrounds, calc, shadows.blur}
\pgfplotsset{compat=1.18}

\definecolor{figGray}{HTML}{7F8C8D}
\definecolor{figBlack}{HTML}{1A1A1A}

\begin{document}

\begin{center}

\rule{\textwidth}{0.8pt}

\vspace{0.65cm}

{\fontsize{16}{20}\selectfont \textbf{Knowledge Graph-Driven Expert-Level \\[4pt] Reasoning for Neuroscience}}

\vspace{0.65cm}

\rule{\textwidth}{0.8pt}

\vspace{0.55cm}

{\fontsize{11}{13}\selectfont
\textbf{Jake Stephen and}
\textbf{Niraj K. Jha}
}

\vspace{0.55cm}

{\fontsize{10}{12}\selectfont
Department of Electrical and Computer Engineering, Princeton University
}

\vspace{1.2cm}

{\fontsize{14}{16}\selectfont \textbf{Abstract}}

\end{center}

\vspace{0.35cm}

\begin{center}
\begin{minipage}{0.82\textwidth}
\fontsize{10}{12.5}\selectfont
Knowledge graph (KG) is an abstraction that can be extracted from text corpora and used for in-depth reasoning. Prior work has leveraged KGs to fine-tune language models (LMs), enabling domain-specific superintelligence.
In this work, we explore whether KG-driven in-depth reasoning capabilities can emerge in neuroscience using only information contained within a single authoritative textbook. The central hypothesis is that structured knowledge, when distilled into a high-quality KG and converted into KG-grounded question-answer (QA) supervision, is sufficient to produce expert-level reasoning through a fine-tuned LM that surpasses large language models (LLMs) in accuracy, while employing orders of magnitude fewer parameters. We construct a textbook-derived KG via a dual-LLM validation pipeline, expand it with a masked LM trained on the KG topology, generate multi-hop QA items, which include QA pairs and reasoning traces, to fine-tune an LM exclusively on KG-derived supervision, and apply reinforcement learning using path-derived KG signals as implicit reward models. Our results demonstrate that deep, mechanistic neuroscience understanding can be induced in the model without reliance on large, heterogeneous web-scale corpora. The KG-based synthetic neuroscience curriculum that readers can quiz themselves on, and the fine-tuned LM, are available at the following GitHub location: \url{https://kg-bottom-up-superintelligence.github.io/neuro-bench}.
\end{minipage}
\end{center}

\setstretch{1.15}
\pagestyle{plain}

\section{Introduction}
\label{sec:intro}

Language models (LMs) traditionally utilized for cross-domain generalization in natural language understanding and generation have recently demonstrated remarkable task-specific reasoning capabilities \cite{anthropic2025claude4,brown2020language, openai2024gpt4o}. Much of this progress has been driven by scale and advanced post-training techniques, enabling models to achieve near-expert performance in highly structured verifiable domains, such as mathematics and programming \cite{deepseek2025r1, snell2024scaling}. However, this top-down training approach, which relies on predicting the next token across broad, general text corpora, often falls short when acquiring the deep, domain-specific abstractions required for specialized scientific fields like neuroscience \cite{dedhia2025bottom}. Achieving domain-specific superintelligence---where an artificial system consistently exceeds the best human experts in a narrow discipline---requires more than surface-level pattern matching; it necessitates a foundational, axiomatic understanding of the domain \cite{bostrom2014superintelligence, burns2023weak}. 

Acquiring deep expertise requires a bottom-up approach, explicitly learning to stitch simple concepts into more complex, higher-order reasoning paths \cite{fodor1975language, kamp1995compositional}. Knowledge graphs (KGs) provide a natural neurosymbolic scaffold for this abstraction, capturing domain primitives as explicit head-relation-tail triples \cite{garcez2023neurosymbolic, pan2023unifying}. A sequence of these triples forms a multi-hop KG path, representing a complex, verifiable concept. Unfortunately, traditional methods for extracting reliable KGs from unstructured domain text are notoriously difficult to scale \cite{zhong2023knowledge}. Conversely, prompting off-the-shelf generative LLMs to perform this extraction often introduces hallucinations and ontological inconsistencies, rendering the resulting KG unsuitable for high-stakes scientific reasoning \cite{belova2025graphmert,ji2023survey}.

Furthermore, teaching a model to perform compositional reasoning over these primitives remains an open challenge. Existing alignment methods, like reinforcement learning from human feedback (RLHF) and direct preference optimization (DPO), optimize models based on final-answer correctness or human preference \cite{ouyang2022training, rafailov2023direct}. In specialized domains, this often leads to reward hacking, where the model relies on superficial correlations rather than a robust, step-by-step verifiable process \cite{shrivastava2025}. To bridge this gap, we must provide models with verifiable, scalable, and grounded supervision that encourages the composition of intermediate axioms, i.e., triples~\cite{kansal2026knowledge}.

To address these challenges, we use an end-to-end pipeline that synthesizes training tasks directly from domain-specific primitives, explicitly enabling the model to acquire and compose these facts for complex multi-hop reasoning. 
Our methodology operates in two distinct phases: graph distillation and curriculum learning.

First, we parse authoritative unstructured texts---specifically focusing on foundational neuroscience via Kandel et al.~\cite{kandel2013principles}---and extract an initial set of relationships using a dual-LLM filtering strategy to form a seed KG. To efficiently scale this graph without introducing generative hallucinations, we use GraphMERT \cite{belova2025graphmert}, an efficient graphical encoder that distills high-quality KGs by jointly learning semantic and syntactic representations. The expanded KG is then passed through a strict re-validation filter to ensure ontological fidelity. 

In the second phase, we traverse multi-hop paths within the validated KG to systematically generate a curriculum of multiple-choice questions (MCQs) and chain-of-thought (CoT) reasoning traces, ranging from one to five hops in complexity. We initially apply supervised fine-tuning (SFT) on the 1-hop and 2-hop question-answer (QA) items (that include the reasoning traces) to ground the model in the domain's axiomatic facts. Subsequently, we employ reinforcement learning (RL) on 2-hop paths. Crucially, rather than relying on human preference, we use the underlying KG paths as an implicit, verifiable reward model \cite{kansal2026knowledge}. By deriving reward signals directly from KG path traversal, we successfully incentivize the model to learn deep compositional reasoning. 

Our core contributions are as follows:
\begin{itemize}
    \item We adapt the scalable, encoder-only distillation approach of Belova et al.~\cite{belova2025graphmert} to a new scientific domain, extracting a highly reliable neuroscience KG from a single authoritative textbook and validating its fidelity through a dual-LLM consensus filter.
    \item Building on the bottom-up curriculum framework of Dedhia et al.~\cite{dedhia2025bottom}, we construct a comprehensive neuroscience QA curriculum spanning 1-hop to 5-hop reasoning, grounded entirely in verifiable KG paths.
    \item Extending the KG-path-rewarded RL paradigm of Kansal and Jha~\cite{kansal2026knowledge}, we demonstrate that a 14B-parameter model further fine-tuned on 2-hop RL generalizes compositional reasoning to 3-to-5-hop queries, surpassing a frontier generalist model within the neuroscience domain.
\end{itemize}

The article is organized as follows.  Section \ref{sec:background} provides background material and discusses related work. Section \ref{sec:theory} discusses the theoretical framework. Section \ref{sec:methodology} provides the methodological path from a textbook to in-depth reasoning. Section \ref{sec:expsetup} discusses the experimental setup and Section \ref{sec:results} provides experimental results. Section \ref{sec:discussion} provides a discussion and limitations. Finally, Section \ref{sec:conclusion} concludes the article.

\section{Background and Related Work}
\label{sec:background}

To contextualize our bottom-up pipeline for in-depth compositional reasoning, we briefly review prior literature across four intersecting domains: language modeling in scientific reasoning, neurosymbolic artificial intelligence, reward modeling in RL, and inference-time scaling.

\subsection{Large Language Models in Scientific Domains}
The scaling of autoregressive transformer architectures has led to profound capabilities in natural language understanding \cite{brown2020language, touvron2023llama}. In specialized domains, efforts to build expert systems have largely relied on continual pre-training over scientific corpora, yielding models such as Galactica for general science \cite{taylor2022galactica} and Med-PaLM for healthcare \cite{singhal2023large}. While these models exhibit robust semantic retrieval, they frequently falter on tasks requiring rigorous compositional reasoning, often suffering from hallucinations and inverse scaling on multi-hop logical deductions \cite{huang2023survey}. Standard top-down pre-training optimizes for distributional matching of text rather than the acquisition of axiomatic truth, limiting their capacity to act as autonomous scientific agents \cite{dedhia2025bottom}.

\subsection{Neurosymbolic AI and Knowledge Graphs}
Neurosymbolic AI seeks to combine the flexible pattern recognition of deep neural networks with the rigorous, interpretable logic of symbolic systems \cite{garcez2023neurosymbolic, sarker2021neuro}. KGs are a widely adopted symbolic representation, encoding facts as head-relation-tail triples. Historically, KG representation learning relied on translational distance models like TransE \cite{bordes2013translating}, graph representation methods \cite{hamilton2017representation}, or graph convolutional networks \cite{kipf2016semi}. More recently, models like COMET \cite{bosselut2019comet} and KEPLER \cite{wang2021kepler} have attempted to fuse language models (LMs) directly with KGs.
However, constructing these domain-specific KGs from unstructured text remains a bottleneck. While generative LLMs can be prompted to extract triples, they are highly sensitive to prompt formulation and prone to ontological hallucinations \cite{ji2023survey, xu2024hallucination}. To bypass these inefficiencies, our pipeline uses GraphMERT \cite{belova2025graphmert}, a compact, encoder-only distillation architecture that extracts reliable KGs by jointly mapping syntactic dependencies and semantic embeddings, ensuring the factual fidelity of the extracted neuroscience axioms.

\subsection{Alignment, RLHF, and Process Supervision}
Aligning LMs with complex tasks has traditionally relied on RLHF based on algorithms like proximal policy optimization \cite{schulman2017proximal, ouyang2022training} or contrastive methods like DPO \cite{rafailov2023direct}. These techniques generally optimize an outcome reward model (ORM) based on the final answer's correctness or human preference. In complex scientific reasoning, ORMs are susceptible to reward hacking, where the model learns superficial correlations rather than robust reasoning steps \cite{gao2023scaling}.
To mitigate this, recent literature has explored process reward models (PRMs) that provide step-by-step supervision \cite{cobbe2021training, lightman2023let}. Yet, training PRMs requires prohibitively expensive human annotations of reasoning trajectories. Our work sidesteps this bottleneck by leveraging the explicit topological structure of the extracted KG. As demonstrated in \cite{kansal2026knowledge}, paths within a validated KG can act as implicit, verifiable reward models. By rewarding models for successfully traversing these multi-hop paths during RL, we provide dense process supervision without human-in-the-loop annotation.

\subsection{Inference-Time Scaling and Curriculum Learning}
Recent paradigms have shifted focus from training-time scale to inference-time compute. Prompting strategies, such as CoT \cite{wei2022chain} and tree-of-thoughts \cite{yao2024tree}, elicit deeper reasoning by forcing models to unfold intermediate steps. Furthermore, inference-time scaling—allocating additional compute during generation to verify or explore alternative reasoning paths—has proven highly effective for complex inference \cite{deepseek2025r1,snell2024scaling}.
However, inference-time techniques cannot conjure domain knowledge that was not internalized during training. To bridge this gap, we adopt a curriculum learning approach \cite{bengio2009curriculum}. By organizing the KG-derived QA tasks from atomic facts (1-hop) to compositional structures (2-hop)~\cite{dedhia2025bottom}, the model progressively composes simple axioms into deep domain expertise, enabling generalization to complex 3-to-5-hop reasoning at inference time.

\section{Basic Framework}
\label{sec:theory}

To formally ground our methodology, we represent neuroscience domain knowledge as a directed, multi-relational KG. A KG is denoted by $\mathcal{G} = (\mathcal{V}, \mathcal{E}, \mathcal{R})$ where $\mathcal{V}$ is the set of entity nodes representing biological and chemical constructs (e.g., ``Dopamine,'' ``Basal Ganglia''), $\mathcal{R}$ is the set of directed relation types mapping their interactions (e.g., ``inhibits,'' ``projects\_to''), and $\mathcal{E} \subseteq \mathcal{V} \times \mathcal{R} \times \mathcal{V}$ is the set of factual edges. A fundamental domain axiom is encoded as a triple $t = (e_s, r, e_t)$.

\subsection{Path-Based Composition and Natural Language Translation}
Achieving complex reasoning requires traversing $\mathcal{G}$ to form logical cascades. We define a multi-hop reasoning path $P_k$ of length $k$ as a contiguous sequence of triples:
\begin{equation}
    P_k = (e_0, r_1, e_1), (e_1, r_2, e_2), \dots, (e_{k-1}, r_k, e_k)
\end{equation}
where the target entity of one triple serves as the source entity for the subsequent triple. In the context of neuroscience, a 1-hop path ($k=1$) represents an atomic fact (e.g., $\texttt{Substantia Nigra} \xrightarrow{\,\textit{produces}\,} \texttt{Dopamine}$). A 3-hop path ($k=3$) captures a mechanistic cascade:
\[
\texttt{Substantia Nigra} \;\xrightarrow{\,\textit{produces}\,}\; \texttt{Dopamine} \;\xrightarrow{\,\textit{inhibits}\,}\; \texttt{Striatum} \;\xrightarrow{\,\textit{regulates}\,}\; \texttt{Motor Cortex}
\]
To bridge the symbolic graph and the autoregressive LM, we define a mapping function $\mathcal{F}: P_k \rightarrow \mathcal{L}_{CoT}$, which translates the structured path $P_k$ into a natural language CoT trace. This trace acts as the ground-truth reasoning trajectory during our supervised training phases \cite{dedhia2025bottom}.

\subsection{Reinforcement Learning with Path-Derived Rewards}
Standard RLHF often optimizes for superficial answer correctness, leading to reward hacking on complex multi-hop queries \cite{shrivastava2025}. Following the paradigm established in \cite{kansal2026knowledge}, we bypass external reward models by using the verifiable topology of $\mathcal{G}$ as an implicit reward signal.

We optimize our policy $\pi_\theta$ using group relative policy optimization (GRPO) \cite{deepseek2025r1,shao2024deepseekmath}. For a given multi-hop question $q$, the model generates a group of $N$ distinct outputs $\{O_1, O_2, \dots, O_N\}$. Each output $O_i$ consists of an internal reasoning trace generated within \texttt{<think>} tags, followed by a final multiple-choice answer enclosed in \texttt{<answer>} tags.

The total reward $R(O_i)$ for a single generation is composed of two additive signals:
\begin{equation}
    R(O_i) = R_{\text{correct}}(O_i) + R_{\text{path}}(O_i)
\end{equation}

The \textbf{correctness reward}, $R_{\text{correct}}(O_i)$, assigns $+1.0$ for a correct final answer and $-1.0$ otherwise, providing a symmetric training signal with a total swing of $2.0$. An additional linear length penalty is subtracted when a completion exceeds a soft token threshold, ramping to a maximum of $-1.0$ at the hard token cap. This provides an explicit gradient against degenerate length explosion, a common failure mode where the policy maximizes completion length to incidentally mention path-relevant tokens.

The \textbf{path alignment reward}, $R_{\text{path}}(O_i)$, provides dense process supervision by measuring the overlap between the model's reasoning trace and the ground-truth KG path. Crucially, this reward is \textit{gated} on answer correctness: $R_{\text{path}}(O_i) = 0$ whenever the predicted answer is incorrect. This gating prevents the model from receiving partial credit for mentioning domain concepts while arriving at wrong conclusions. When the gate is satisfied, we parse the \texttt{<think>} block, normalize its tokens, and compute:
\begin{equation}
    R_{\text{path}}(O_i) = \min\!\Big(\big(0.8 \cdot \text{coverage}(\mathcal{T}_i, \mathcal{P}) + 0.3 \cdot \mathbbm{1}[|\mathcal{T}_i \cap \mathcal{P}| \geq 2]\big) \cdot \rho(\mathcal{T}_i),\; 0.8\Big)
    \label{eq:Rpath}
\end{equation}
where $\text{coverage}(\mathcal{T}_i, \mathcal{P}) = |\mathcal{T}_i \cap \mathcal{P}| / |\mathcal{P}|$ measures the fraction of ground-truth path concepts mentioned in the reasoning trace, the indicator term requires at least two distinct path hits, and $\rho(\mathcal{T}_i) \in [0, 1]$ is a repetition penalty that down-weights outputs with degenerate token repetition. The cap at $0.8$ ensures that $R_{\text{correct}}$ remains the dominant training signal, preventing the policy from optimizing path overlap at the expense of answer accuracy.

GRPO computes the advantage $A_i$ by normalizing rewards relative to the generated group, removing the need for a separate value network:
\begin{equation}
    A_i = \frac{R(O_i) - \mu_R}{\sigma_R}
\end{equation}
where $\mu_R$ and $\sigma_R$ are the mean and standard deviation of the rewards across the $N$ generations. A frozen copy of the SFT model serves as the reference policy $\pi_{\text{ref}}$. A Kullback-Leibler (KL) divergence penalty anchors the active policy to prevent entropy collapse. The policy is updated by maximizing the clipped surrogate objective:
\begin{equation}
    \mathcal{L}_{\text{GRPO}}(\theta) = -\mathbb{E}_{q,\{O_i\}} \left[ \frac{1}{N} \sum_{i=1}^{N} \min\!\left( r_i(\theta)\, A_i,\; \text{clip}(r_i(\theta), 1{-}\epsilon, 1{+}\epsilon)\, A_i \right) - \beta\, D_{\text{KL}}\!\left[\pi_\theta \| \pi_{\text{ref}}\right] \right]
\end{equation}
where $r_i(\theta) = \pi_\theta(O_i|q) / \pi_{\text{ref}}(O_i|q)$ is the importance ratio. By jointly enforcing answer correctness, path-grounded reasoning, length discipline, and proximity to the SFT prior, the reward design forces the model to acquire robust, compositional reasoning.

\section{Methodology: Textbook to In-depth Compositional Reasoning}
\label{sec:methodology}

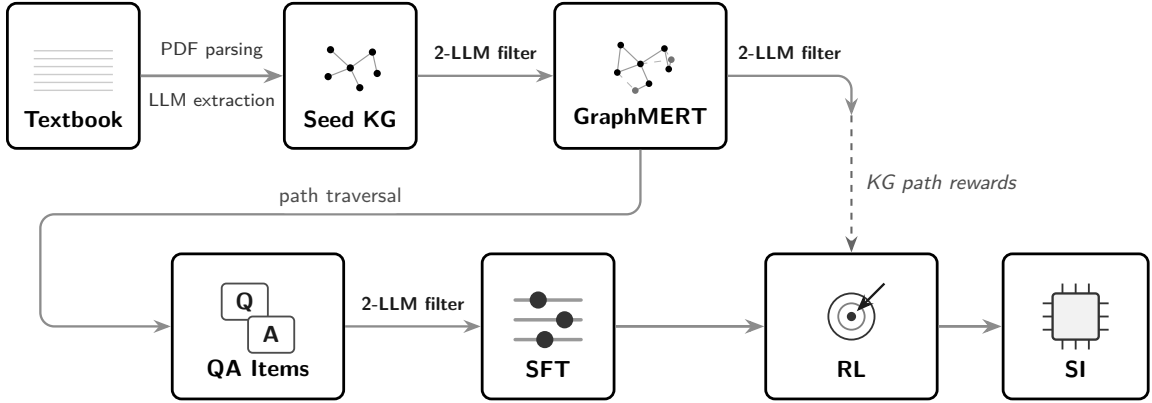
\begin{figure}[htbp]
    \centering
    \resizebox{\textwidth}{!}{

\begin{tikzpicture}[
  >=Stealth, font=\sffamily,
  box/.style={draw=black,fill=white,rounded corners=4pt,
    minimum height=2.2cm,minimum width=2.0cm,line width=1.2pt,align=center},
  wbox/.style={box,minimum width=2.6cm},
  arr/.style={->,line width=1.0pt,color=black!45},
  bigarr/.style={->,line width=1.2pt,color=black!45},
]

\def\ry{-3.8}

\node[box]  (book) at (0,0) {};
\node[box]  (skg)  at (4.2,0) {};
\node[wbox] (gm)   at (8.6,0) {};

\node[wbox] (qac)  at (2.8,\ry) {};
\node[box]  (sft)  at (7.2,\ry) {};
\node[wbox] (rl)   at (11.8,\ry) {};
\node[box,minimum width=2.2cm,minimum height=2.2cm,rounded corners=4pt] (si) at (15.2,\ry) {};

\begin{scope}[shift={(book.center)}]
  \foreach \y in {0.38,0.26,0.14,0.02,-0.1,-0.22}
    {\draw[black!20,line width=0.6pt] (-0.6,\y) -- (0.6,\y);}
  \node[font=\sffamily\bfseries\small,text=black] at (0,-0.65) {Textbook};
\end{scope}

\begin{scope}[shift={(skg.center)}]
  \begin{scope}[shift={(0,0.12)},scale=1.2]
    \foreach \a/\b/\c/\d in {0/0/0.28/0.2, 0/0/-0.25/0.23, 0/0/0.12/-0.25,
      0/0/-0.28/-0.11, 0.28/0.2/0.34/-0.07}{
      \draw[black!40,line width=0.5pt] (\a,\b)--(\c,\d);}
    \foreach \a/\b in {0/0,0.28/0.2,-0.25/0.23,0.12/-0.25,-0.28/-0.11,0.34/-0.07}{
      \fill[black] (\a,\b) circle(1.3pt);}
  \end{scope}
  \node[font=\sffamily\bfseries\small,text=black] at (0,-0.65) {Seed KG};
\end{scope}

\begin{scope}[shift={(gm.center)}]
  \begin{scope}[shift={(0,0.18)},scale=1.3]
    \foreach \a/\b/\c/\d in {0/0/0.28/0.18, 0/0/-0.23/0.22, 0/0/0.1/-0.23,
      0/0/-0.27/-0.1, 0.28/0.18/0.33/-0.06, -0.23/0.22/-0.27/-0.1,
      0.1/-0.23/-0.06/-0.31, 0.33/-0.06/0.36/0.05}{
      \draw[black!40,line width=0.5pt] (\a,\b)--(\c,\d);}
    \draw[black!30,dashed,line width=0.4pt] (-0.27,-0.1)--(-0.06,-0.31);
    \draw[black!30,dashed,line width=0.4pt] (0.28,0.18)--(0.36,0.05);
    \draw[black!30,dashed,line width=0.4pt] (0,0)--(0.36,0.05);
    \foreach \a/\b in {0/0,0.28/0.18,-0.23/0.22,0.1/-0.23,-0.27/-0.1,0.33/-0.06}{
      \fill[black] (\a,\b) circle(1.1pt);}
    \foreach \a/\b in {-0.06/-0.31,0.36/0.05}{
      \fill[black!55] (\a,\b) circle(1.1pt);}
  \end{scope}
  \node[font=\sffamily\bfseries\small,text=black] at (0,-0.65) {GraphMERT};
\end{scope}

\begin{scope}[shift={(qac.center)}]
  \draw[black!70, fill=white, line width=0.8pt, rounded corners=2pt] (-0.55, 0.1) rectangle (0.15, 0.65);
  \draw[black!70, fill=white, line width=0.8pt, rounded corners=2pt] (-0.15, -0.4) rectangle (0.55, 0.15);
  \node[font=\sffamily\small\bfseries, text=black!90] at (-0.2, 0.37) {Q};
  \node[font=\sffamily\small\bfseries, text=black!90] at (0.2, -0.12) {A};
  \node[font=\sffamily\bfseries\small,text=black] at (0,-0.65) {QA Items};
\end{scope}

\begin{scope}[shift={(sft.center)}]
  \draw[black!40, line width=1.5pt, line cap=round] (-0.5, 0.4) -- (0.5, 0.4);
  \draw[black!40, line width=1.5pt, line cap=round] (-0.5, 0.1) -- (0.5, 0.1);
  \draw[black!40, line width=1.5pt, line cap=round] (-0.5, -0.2) -- (0.5, -0.2);
  \fill[black!80] (-0.15, 0.4) circle (0.13);
  \fill[black!80] (0.25, 0.1) circle (0.13);
  \fill[black!80] (-0.05, -0.2) circle (0.13);
  \node[font=\sffamily\bfseries\small,text=black] at (0,-0.65) {SFT};
\end{scope}

\begin{scope}[shift={(rl.center)}]
  \draw[black!80, line width=0.8pt] (0,0.15) circle (0.35);
  \draw[black!50, line width=0.8pt] (0,0.15) circle (0.2);
  \fill[black!80] (0,0.15) circle (0.07);
  \draw[->, >=Stealth, black!90, line width=1pt] (0.5, 0.65) -- (0.08, 0.23);
  \node[font=\sffamily\bfseries\small,text=black] at (0,-0.65) {RL};
\end{scope}

\begin{scope}[shift={(si.center)}]
  \draw[black!80, fill=black!5, line width=1pt, rounded corners=1.5pt] (-0.35, -0.2) rectangle (0.35, 0.5);
  \foreach \x in {-0.2, 0, 0.2} {
    \draw[black!80, line width=0.8pt] (\x, 0.5) -- (\x, 0.65);
    \draw[black!80, line width=0.8pt] (\x, -0.2) -- (\x, -0.35);
  }
  \foreach \y in {-0.05, 0.15, 0.35} {
    \draw[black!80, line width=0.8pt] (-0.35, \y) -- (-0.5, \y);
    \draw[black!80, line width=0.8pt] (0.35, \y) -- (0.5, \y);
  }
  \node[font=\sffamily\bfseries\small,text=black] at (0,-0.65) {SI};
\end{scope}


\draw[bigarr] (book.east) -- node[above,font=\sffamily\scriptsize,text=black!80,yshift=3pt]{PDF parsing}
                             node[below,font=\sffamily\scriptsize,text=black!80,yshift=-3pt]{LLM extraction} (skg.west);

\draw[arr] (skg.east) -- node[above,font=\sffamily\bfseries\scriptsize,text=black!90,yshift=3pt]{2-LLM filter} (gm.west);

\draw[arr, rounded corners=6pt] (gm.south) 
    -- (8.6, -2.1) 
    -- node[above,font=\sffamily\footnotesize,text=black!70,fill=white,inner sep=3pt]{path traversal} (-0.5, -2.1)
    -- (-0.5, \ry) 
    -- (qac.west);

\draw[arr, rounded corners=6pt] (gm.east) -- node[above,font=\sffamily\bfseries\scriptsize,text=black!90,yshift=3pt]{2-LLM filter} (11.8, 0) -- (11.8, -0.6);
\draw[->, dashed, line width=0.9pt, color=black!60, rounded corners=6pt] (11.8, -0.6) 
    -- node[right,font=\sffamily\footnotesize\itshape,text=black!70,xshift=2pt]{KG path rewards} (rl.north);

\draw[arr] (qac.east) -- node[above,font=\sffamily\bfseries\scriptsize,text=black!90,yshift=3pt]{2-LLM filter} (sft.west);

\draw[bigarr] (sft.east) -- (rl.west);

\draw[bigarr] (rl.east) -- (si.west);

\end{tikzpicture}}
    \caption{End-to-end pipeline for bottom-up 
    domain-specific superintelligence (SI).}
    \label{fig:pipeline}
\end{figure}

Our end-to-end pipeline consists of five primary phases: (1) seed KG extraction from textbook, (2) KG validation and GraphMERT expansion, (3) QA item generation, (4) SFT, and (5) RL. Fig.~\ref{fig:pipeline} gives a high-level overview.

\subsection{Phase 1: Textbook Parsing and Seed KG Extraction}

To construct the seed KG, we parsed the digitized PDF of \cite{kandel2013principles} into structured text units. This applies a sliding-window chunker to produce overlapping text segments of approximately 300 tokens. Each text unit is independently processed by an open-weight Qwen3-14B LM~\cite{bai2023qwen} to extract typed entities and directed relational triples.

The extraction prompt (reproduced in full in Appendix~\ref{app:prompt}) follows a structured few-shot format. The system message instructs the model to identify entities belonging to one of six biomedical ontological categories---\textsc{Anatomical Structure}, \textsc{Molecular Entity}, \textsc{Cellular Component}, \textsc{Process}, \textsc{Clinical Entity}, and \textsc{Conceptual Entity}---and extract directed relationships drawn exclusively from a closed vocabulary of 40 neuroscience-specific relation types (enumerated in Appendix~\ref{app:relations}). A single worked example grounding the prompt in neuroscience terminology accompanies every inference call as a one-shot demonstration.

\subsection{Phase 2: The Two-LLM Validation Strategy}
LLM-generated KGs inherently suffer from hallucinations and varying granularity. To ensure the highest fidelity, we instituted a ``two-LLM validation'' strategy. Every extracted triple $t$ is evaluated by two distinct open-weight LLM families (gpt-oss-20b and Mistral-Nemo-12B). 

The validation prompt requires the LLMs to return a reasoning trace alongside a definitive ``Yes'' or ``No'' regarding the factual accuracy of the triple in the context of the textbook. A triple is preserved in the seed KG only if both models reach a consensus of ``Yes''. Out of an initial batch, this process successfully filters out 1,843 faulty relationships, preserving 6,157 high-confidence triples.

\subsection{Phase 3: GraphMERT Training and KG Expansion}
To expand our seed KG beyond explicitly stated textual facts to implicit, mechanistic derivations, we use GraphMERT \cite{belova2025graphmert}. 

\noindent
\textbf{Preprocessing and tokenization:} The seed KG is transformed to match GraphMERT's schema. 

\noindent
\textbf{Masked language model training:} We configure GraphMERT with six layers, 12 attention heads, 512 root nodes, and a maximum of 2048 nodes. To accommodate the semantic space of neuroscience triples, we set the leaf count parameter to three tokens per root. Training is executed using a batch size $B$ of 8 over 20 epochs, optimized via an initial learning rate grid search.

\noindent
\textbf{Tail prediction and re-validation:} The model successfully proposed thousands of new relations based on the learned graph topology. These new triples underwent the same two-LLM validation process as Phase 2. The combination of the validated seed KG and the GraphMERT expansion yields a highly dense, comprehensive neuroscience KG containing:
\begin{itemize}
    \item \textbf{Total nodes:} 9,187
    \item \textbf{Total triples:} 19,755
    \item \textbf{Average node degree:} 2.15
\end{itemize}

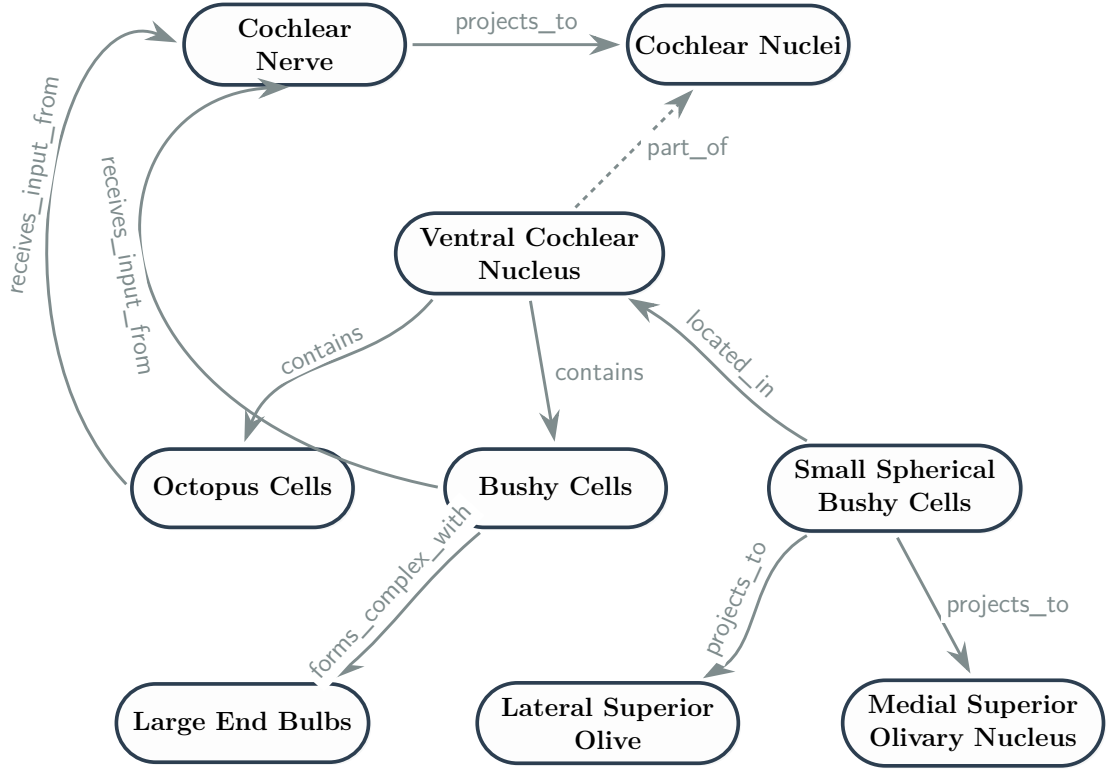
\begin{figure}[!ht]
    \centering
    \resizebox{\textwidth}{!}{%
\providecolor{borderCol}{HTML}{2C3E50}
\providecolor{nodeFill}{HTML}{FDFDFD}
\providecolor{shadowCol}{HTML}{BDC3C7}
\begin{tikzpicture}[
    >={Stealth[length=6mm, width=4mm]},
    kgNode/.style={
        draw=borderCol,
        line width=2pt,
        fill=nodeFill,
        rounded rectangle,
        rounded rectangle west arc=convex,
        rounded rectangle east arc=convex,
        align=center,
        minimum height=1.5cm,
        minimum width=4.8cm,
        inner sep=8pt,
        font=\rmfamily\bfseries\Large,
        drop shadow={opacity=0.15, shadow xshift=1.5pt, shadow yshift=-1.5pt, fill=shadowCol}
    },
    kgLabel/.style={
        fill=white,
        inner sep=3pt,
        font=\sffamily\Large\color{figGray},
        midway, sloped, above=1pt
    },
    kgLabelBelow/.style={
        fill=white,
        inner sep=3pt,
        font=\sffamily\Large\color{figGray},
        midway, sloped, below=1pt
    },
    kgLabelStraight/.style={
        fill=white,
        inner sep=3pt,
        font=\sffamily\Large\color{figGray},
        midway, above=2pt
    },
    liMain/.style={
        ->, draw=figGray, line width=1.6pt,
        shorten >=3pt, shorten <=3pt
    },
    liStructural/.style={
        ->, draw=figGray, line width=1.6pt, dashed,
        shorten >=3pt, shorten <=3pt
    }
]
    \node[kgNode] (CN)   at (0, 0)          {Cochlear\\Nerve};
    \node[kgNode] (CNuc) at (8.5, 0)        {Cochlear Nuclei};

    \node[kgNode] (VCN)  at (4.5, -4)       {Ventral Cochlear\\Nucleus};

    \node[kgNode] (OC)   at (-1, -8.5)      {Octopus Cells};
    \node[kgNode] (BC)   at (5, -8.5)       {Bushy Cells};
    \node[kgNode] (SSBC) at (11.5, -8.5)    {Small Spherical\\Bushy Cells};

    \node[kgNode] (LEB)  at (-1, -13)       {Large End Bulbs};
    \node[kgNode] (LSO)  at (6, -13)        {Lateral Superior\\Olive};
    \node[kgNode] (MSO)  at (13, -13)       {Medial Superior\\Olivary Nucleus};

    \draw[liMain] (CN) -- node[kgLabelStraight] {projects\_to} (CNuc);

    \draw[liStructural] (VCN) -- node[kgLabel, sloped=false, right=3pt] {part\_of} (CNuc);

    \draw[liMain] (VCN.south west) to[out=-130, in=70]
        node[kgLabel] {contains} (OC.north);

    \draw[liMain] (VCN.south) -- node[kgLabel, sloped=false, right=3pt] {contains} (BC.north);

    \draw[liMain] (SSBC.north west) to[out=150, in=-30]
        node[kgLabel] {located\_in} (VCN.south east);

    \draw[liMain] (BC.south west) to[out=-140, in=40]
        node[kgLabel] {forms\_complex\_with} (LEB.north east);

    \draw[liMain] (SSBC.south west) to[out=-150, in=30]
        node[kgLabel] {projects\_to} (LSO.north east);

    \draw[liMain] (SSBC.south) -- node[kgLabel, sloped=false, right=3pt] {projects\_to} (MSO.north);

    \draw[liMain] (BC.west) .. controls (-4.5, -7) and (-4, -1) ..
        node[kgLabelBelow] {receives\_input\_from} (CN.south);

    \draw[liMain] (OC.west) .. controls (-6, -5.5) and (-4.5, 1) ..
        node[kgLabel] {receives\_input\_from} (CN.west);

\end{tikzpicture}%
    }
    \caption{Sample subset of the neuroscience KG}
    \label{fig:kg_subset}
\end{figure}

A small subset of the neuroscience KG thus derived is shown in Fig.~\ref{fig:kg_subset}.

\subsection{Phase 4: Multi-hop QA Curriculum Generation}

With the finalized KG ($\sim$20k triples), we computed an adjacency list and ran a depth-first path traversal to extract multi-hop causal pathways of 1-5 hops in length. To control combinatorial explosion and remove uninformative paths, we applied three pruning strategies: (1) \textbf{hub node removal}, where the top 1\% of nodes by degree were penalized to prevent paths from routing through overly generic entities; (2) \textbf{relation filtering}, which downweighted semantically weak edges (e.g., \texttt{associated\_with}, \texttt{located\_in}, \texttt{part\_of}) that add structural noise without causal specificity; and (3) \textbf{transitive redundancy pruning}, which discarded multi-hop paths whose endpoints were already directly connected, ensuring the curriculum emphasizes genuinely indirect reasoning chains. 

We converted sampled paths into natural-language MCQs paired with step-by-step CoT reasoning traces. Initial QA item generation was performed using Gemini 2.5 Flash, with all subsequent validation and refinement steps handled by Gemini 2.0 Flash. Each item was further validated through cross-checking with two distinct LLM families to ensure factual accuracy and reasoning coherence. The final QA curriculum comprises:
\begin{itemize}
    \item 1-hop: all 13,919 items (exhaustive coverage of seed KG edges)
    \item 2-hop: 30,000 items (of which 5,000 reserved for RL)
    \item 3-to-5-hop: 1,000 items for each hop length (for evaluation)
\end{itemize}

\subsection{Phase 5: Supervised Fine-Tuning and Reinforcement Learning}
\label{subsec:rl_sft}

\textbf{SFT phase:}
We perform SFT on the base Qwen3-14B model using the 1-hop and 2-hop QA items, instilling the domain's axiomatic facts and aligning the model to the structured \texttt{<think>}/\texttt{<answer>} CoT format required for subsequent RL. Training uses low-rank adaptation (LoRA) modules~\cite{hu2021lora}, which are subsequently merged into the base weights to produce a single set of full-precision parameters before the RL phase begins.

\noindent
\textbf{RL phase:}
Starting from the merged SFT checkpoint, we apply GRPO~\cite{shao2024deepseekmath} using 5,000 held-out 2-hop QA item prompts. The reward formulation and GRPO objective were described in full in Section~\ref{sec:theory}; we describe the implementation details here.

A frozen copy of the SFT checkpoint serves as the reference policy $\pi_{\text{ref}}$, with KL penalty $\beta = 0.12$ to prevent entropy collapse. We generate $N{=}4$ completions per prompt and parse each output's \texttt{<think>} block against the ground-truth KG path to compute $R_{\text{path}}$, gated on answer correctness. 

We perform full-parameter fine-tuning (no LoRA) using DeepSpeed ZeRO Stage-3 with optimizer CPU offloading across 4$\times$~A100-80GB GPUs, with gradient checkpointing enabled to reduce activation memory. Table~\ref{tab:rl_hyperparams} summarizes all training hyperparameters.

\begin{table}[htbp]
\centering
\caption{GRPO training hyperparameters.}
\label{tab:rl_hyperparams}
\begin{tabular}{ll}
\toprule
\textbf{Hyperparameter} & \textbf{Value} \\
\midrule
Learning rate & $2 \times 10^{-6}$ \\
KL penalty ($\beta$) & 0.12 \\
Generations per prompt ($N$) & 4 \\
Max completion length & 1792 tokens \\
Gradient accumulation steps & 16 \\
Per-device batch size & 1 \\
Optimizer & AdamW (CPU offload) \\
Precision & bfloat16 \\
Learning rate schedule & Constant with 5\% warmup \\
Max gradient norm & 1.0 \\
Sampling temperature & 0.6 \\
Top-$p$ & 0.9 \\
Training epochs & 3 \\
\bottomrule
\end{tabular}
\end{table}

\begin{algorithm}[t]
\caption{GRPO Training with KG Implicit Rewards (adapted from~\cite{kansal2026knowledge})}
\label{alg:rl_loop}
\begin{algorithmic}[1]
\Require Policy $\pi_\theta$, Reference policy $\pi_{\text{ref}}$, Knowledge graph $G$, Prompts $Q$, KL coefficient $\beta$
\State Initialize $\pi_\theta \gets \pi_{\text{SFT}}$, \quad $\pi_{\text{ref}} \gets \pi_{\text{SFT}}$ (frozen)
\For{each training step}
    \State Sample batch of questions $\{q_1, \dots, q_B\} \sim Q$
    \For{each $q_j$}
        \State Generate $N$ completions $\{O_1, \dots, O_N\} \sim \pi_\theta(\cdot \mid q_j)$
        \For{each $O_i$}
            \State Parse reasoning trace $\hat{T}_i$ from \texttt{<think>} tags
            \State Parse predicted answer $\hat{A}_i$ from \texttt{<answer>} tags
            \State $R_{\text{correct}}(O_i) \gets \begin{cases} +1.0 & \text{if } \hat{A}_i = A^* \\ -1.0 & \text{otherwise} \end{cases}$ \; $- \; \min\!\Big(\frac{\max(0,\;\text{len}(O_i) - L_{\text{soft}})}{L_{\max} - L_{\text{soft}}},\; 1.0\Big)$
            \State $R_{\text{path}}(O_i) \gets \min\big((0.8 \cdot \text{cov}(\hat{T}_i, G) + 0.3 \cdot \mathbbm{1}[\text{hits} \geq 2]) \cdot \rho(\hat{T}_i),\; 0.8\big)$
            \State $R(O_i) \gets R_{\text{correct}}(O_i) + R_{\text{path}}(O_i)$
        \EndFor
        \State Compute group-relative advantages: $A_i \gets \frac{R(O_i) - \text{mean}(\{R(O_k)\}_{k=1}^N)}{\text{std}(\{R(O_k)\}_{k=1}^N)}$
    \EndFor
    \State Update $\theta$ via clipped surrogate objective with KL penalty:
    \State $\theta \gets \theta - \eta \nabla_\theta \Big[ -\frac{1}{BN}\sum_{j,i} \min\!\big(r_i A_i,\; \text{clip}(r_i, 1{-}\epsilon, 1{+}\epsilon) A_i\big) + \beta\, D_{\text{KL}}[\pi_\theta \| \pi_{\text{ref}}] \Big]$
\EndFor
\end{algorithmic}
\end{algorithm}

Algorithm~\ref{alg:rl_loop} details the complete GRPO training loop. At each step, a batch of multi-hop questions is sampled and the active policy generates $N$ candidate completions per question. Each completion is decomposed into its reasoning trace and final answer via structured tag parsing, and the two reward components---correctness and path alignment---are computed independently and summed. Advantages are then normalized within each generation group, eliminating the need for a learned value network. Finally, the policy parameters are updated using the clipped surrogate objective, with a KL divergence term anchoring the active policy to the frozen SFT reference to prevent entropy collapse. This loop iterates over the full set of 2-hop training prompts for three epochs, progressively sharpening the model's compositional reasoning without any human preference annotation.

\section{Experimental Setup}
\label{sec:expsetup}
To rigorously evaluate the induction of compositional reasoning, we establish the
following experimental conditions and baselines.

\subsection{Models Evaluated}

Our primary base architecture is Qwen3-14B. We evaluate the following variants to isolate
the contribution of each training phase:

\begin{enumerate}
    \item \textbf{Base model}: Qwen3-14B (zero-shot, no domain-specific training).
    \item \textbf{SFT-only}: Qwen3-14B fine-tuned on $\sim$14k 1-hop and $\sim$25k 2-hop QA items using LoRA, subsequently merged into full weights.
    \item \textbf{SFT+RL}: The fully realized bottom-up pipeline. The merged SFT checkpoint undergoes full-parameter GRPO training with KG implicit rewards on 5k multi-hop prompts, using the configuration detailed in Table 1.
\end{enumerate}

In addition, we benchmark against a frontier proprietary model, specifically Gemini 3.1
Pro, to determine whether the 14B specialized model can outperform a
generalized multi-trillion-parameter model within the neuroscience domain.

\subsection{Evaluation Metrics}

Models are assessed on the test dataset comprising 3-hop to 5-hop questions. We evaluate them
along two axes:

\begin{itemize}
    \item \textbf{Accuracy (Acc)}: The proportion of questions where the final multiple-choice selection is correct, extracted from the model's \texttt{<answer>} tags.
    \item \textbf{Compositional generalization}: The performance decay rate as question hop-length increases from $k=3$ to $k=5$. An ideal model should maintain stable accuracy across longer reasoning chains, indicating true compositional reasoning rather than pattern matching.
\end{itemize}

\section{Experimental Results}
\label{sec:results}

Our evaluation addresses a central empirical question: Can a 14B parameter LM, trained exclusively on a textbook-derived KG curriculum, match or exceed the domain-specific reasoning capabilities of generalist frontier models? The results, detailed ahead,
affirm this hypothesis across all evaluated conditions. We structure our analysis as follows. We first ground the quantitative metrics with a qualitative inspection of reasoning traces (\S\ref{subsec:qualitative}), then present systematic analyses of overall performance (\S\ref{subsec:overall}), compositional generalization (\S\ref{subsec:compositional}), the role of prior SFT in enabling RL (\S\ref{subsec:sft_enables_rl}), frontier model comparison (\S\ref{subsec:frontier}), and a discussion of observed failure modes (\S\ref{subsec:failures}).

\subsection{Qualitative Analysis: KG-Grounded Reasoning}
\label{subsec:qualitative}

To contextualize the aggregate accuracy metrics, we examine whether the model's internal reasoning traces faithfully traverse the underlying KG paths. Fig.~\ref{fig:qualitative} presents a representative 3-hop example from the evaluation set, showing the ground-truth KG path alongside the model's generated \texttt{<think>} trace. Bolded entities in the reasoning trace correspond to nodes in the ground-truth path, demonstrating that the model internalizes the graph's causal topology rather than relying on surface-level lexical cues.

\begin{figure}[htbp]
\centering
\small
\fbox{\parbox{0.94\textwidth}{%
\textbf{KG Path (3-hop):}\\[4pt]
\texttt{local interneurons}
$\;\xrightarrow{\textit{projects\_to}}\;$
\texttt{cerebellum}
$\;\xrightarrow{\textit{participates\_in}}\;$
\texttt{sensorimotor transformations}
$\;\xrightarrow{\textit{mediates\_signal\_for}}\;$
\texttt{voluntary motor behavior}

\medskip\hrule\medskip

\textbf{Question:} A research team develops a transgenic mouse model where a specific population of small, highly branched neurons, densely distributed within the intermediate gray matter of the spinal cord and specific brainstem reticular formation nuclei, can be reversibly silenced. These neurons are distinguished by their short-range axonal projections, extensive local dendritic trees, and their critical function in integrating convergent sensory afferents and descending modulatory inputs before relaying processed information to larger projection neurons. When these localized neurons are functionally inhibited, the mice exhibit intact muscle strength, normal muscle tone, and unimpaired monosynaptic reflexes. However, their ability to execute fine motor skills, such as accurately grasping small food pellets or precisely adjusting limb trajectory to navigate a dynamic obstacle course, becomes profoundly impaired. They frequently overshoot or undershoot targets, display jerky movements, and struggle to learn novel, intricate motor sequences, despite showing clear intent to perform the actions. Given this presentation, the primary functional deficit observed in these mice most directly impacts:

\medskip

\begin{tabular}{@{}ll@{}}
A. & The accurate gating and filtering of primary somatosensory inputs in the dorsal horn. \\
B. & The rapid adaptation of feedforward motor commands based on predicted sensory \\
   & consequences. \\
C. & The rhythmic pattern generation underlying stereotypical behaviors like locomotion \\
   & and respiration. \\
D. & The integration of proprioceptive feedback for maintaining postural stability against \\
   & gravity.
\end{tabular}

\medskip\hrule\medskip

\textbf{Model Output} \hfill \textit{Answer: B \checkmark}\\[4pt]
\texttt{<think>}\\
The mice have impaired \textbf{fine motor skills} despite intact muscle strength, tone, and reflexes. This suggests a deficit in higher-level \textbf{motor control}, specifically the adaptation of \textbf{motor commands} based on predicted \textbf{sensory consequences}. The mice struggle with precise movements and learning new \textbf{motor sequences}, indicating a problem with adapting motor commands. Option~A is incorrect because the mice's \textbf{somatosensory} processing appears intact, as evidenced by the normal reflexes. Option~C is incorrect because the \textbf{rhythmic pattern generation} for stereotypical behaviors is likely intact, given the normal reflexes and muscle strength. Option~D is incorrect because the integration of \textbf{proprioceptive feedback} for maintaining \textbf{postural stability} is likely intact, given the normal muscle tone and reflexes. Therefore, B is the correct answer.\\
\texttt{</think>}
}}
\caption{Qualitative example of KG-grounded reasoning. The model's \texttt{<think>} trace implicitly traverses the ground-truth path: It identifies the silenced neurons as \textbf{local interneurons} whose disruption impairs \textbf{cerebellar}-dependent \textbf{sensorimotor transformations}, ultimately degrading \textbf{voluntary motor behavior}. Bolded terms denote entities aligned with KG nodes.}
\label{fig:qualitative}
\end{figure}

This trace exhibits a reasoning strategy that is structurally distinct from standard CoT generation in ungrounded models. Each inferential step in the \texttt{<think>} block corresponds to the traversal of a specific edge in the internalized KG: The model identifies the affected neural population (local interneurons), links their projections to a downstream structure (cerebellum), maps that structure to a functional role (sensorimotor transformations), and connects the functional role to a behavioral outcome (voluntary motor behavior). This four-stage cascade mirrors the ground-truth KG path with high fidelity.

The model also demonstrates structured elimination of distractors. Its rejection of options A, C, and D is grounded in explicit reasoning about preserved functional capacities: Intact reflexes rule out somatosensory gating (A), intact muscle strength rules out pattern generation (C), and intact muscle tone rules out proprioceptive integration (D). This pattern of elimination-by-preserved-function is characteristic of expert clinical reasoning in neuroscience~\cite{kandel2013principles} and is consistent with the model having acquired structured diagnostic methodology through the KG-grounded curriculum. A systematic annotation study quantifying path alignment rates across a larger sample of evaluation traces would further validate this observation; we identify this as a direction for future empirical work.

\subsection{Overall Performance and the Impact of Bottom-Up Training}
\label{subsec:overall}

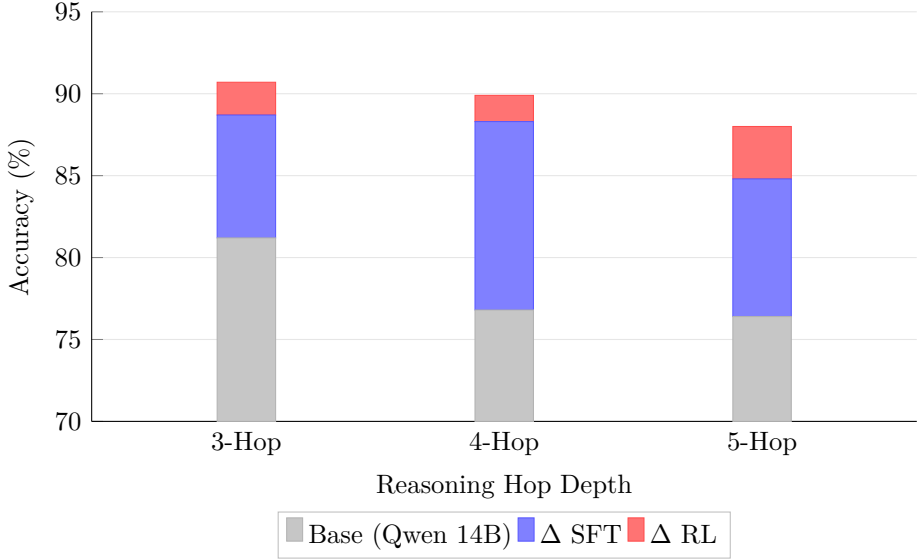
\begin{figure}[htbp]
\centering
\begin{tikzpicture}
    \begin{axis}[
        ybar stacked,
        bar width=22pt,
        width=0.82\textwidth,
        height=7cm,
        ylabel={Accuracy (\%)},
        xlabel={Reasoning Hop Depth},
        symbolic x coords={3-Hop, 4-Hop, 5-Hop},
        xtick=data,
        ymin=70, ymax=95,
        enlarge x limits=0.3,
        legend style={
            at={(0.5,-0.22)},
            anchor=north,
            legend columns=3,
            font=\small,
            draw=gray!50,
        },
        axis lines*=left,
        ymajorgrids=true,
        grid style={gray!20},
        tick label style={font=\small},
        ylabel style={font=\small},
        xlabel style={font=\small},
    ]
    \addplot[fill=gray!45, draw=gray!60] coordinates {
        (3-Hop, 81.2) (4-Hop, 76.8) (5-Hop, 76.4)
    };
    \addplot[fill=blue!50, draw=blue!65] coordinates {
        (3-Hop, 7.5) (4-Hop, 11.5) (5-Hop, 8.4)
    };
    \addplot[fill=red!55, draw=red!70] coordinates {
        (3-Hop, 2.0) (4-Hop, 1.6) (5-Hop, 3.2)
    };
    \legend{Base (Qwen 14B), $\Delta$ SFT, $\Delta$ RL}
    \end{axis}
\end{tikzpicture}
\caption{Ablation of training phase contributions across hop depths, computed from Table~\ref{tab:results_table}. The SFT phase (blue) provides the largest absolute gain at all depths. The RL phase (red) contributes a smaller but disproportionately depth-dependent gain: its contribution at 5-hop (+3.2 pp) exceeds its contribution at 3-hop (+2.0 pp), indicating that path-derived rewards preferentially strengthen compositional reasoning at greater depths.}
\label{fig:ablation_bar}
\end{figure}

The baseline Qwen3-14B model achieves 78.1\% average accuracy across the 3-to-5-hop evaluation set, reflecting neuroscience knowledge encoded through general-purpose pre-training but organized in a form that does not readily support deep compositional reasoning. Table~\ref{tab:results_table} and Fig.~\ref{fig:ablation_bar} decompose the contribution of each training phase.

SFT yields a +9.2 pp average gain (78.1\% $\to$ 87.3\%), with the largest improvement at four hops (+11.5 pp), where the base model exhibits its steepest drop from 3-hop performance. This is consistent with the observation that SFT primarily closes factual coverage gaps: 4-hop chains are long enough to expose missing domain knowledge yet short enough that reasoning coherence is not the binding constraint.

The RL phase adds +2.2 pp on average, but its contribution is distributed asymmetrically: +2.0 pp at three hops, +1.6 pp at four hops, and +3.2 pp at five hops. The largest RL gain occurs at five hops---precisely where SFT leaves the greatest residual gap---indicating that path-derived rewards address a qualitatively different bottleneck than SFT. Where SFT expands the model's factual inventory, RL strengthens its ability to maintain reasoning coherence across extended chains. This complementarity is reflected in the stacked contributions visible in Fig.~\ref{fig:ablation_bar}: SFT provides a roughly uniform lift, whereas RL's contribution grows with depth.

\begin{table}[htbp]
\centering
\caption{Overall performance metrics across models.}
\label{tab:results_table}
\resizebox{\textwidth}{!}{%
\begin{tabular}{lcccc}
\toprule
\textbf{Model} & \textbf{3-hop Acc (\%)} & \textbf{4-hop Acc (\%)} & \textbf{5-hop Acc (\%)} & \textbf{Avg Acc (\%)} \\
\midrule
Gemini 3.1 Pro       & 86.9 & 85.2 & 82.2 & 84.8 \\
Qwen 14B (Base)      & 81.2 & 76.8 & 76.4 & 78.1 \\
Qwen 14B (SFT)       & 88.7 & 88.3 & 84.8 & 87.3 \\
Qwen 14B (SFT + RL)  & \textbf{90.7} & \textbf{89.9} & \textbf{88.0} & \textbf{89.5} \\
\bottomrule
\end{tabular}%
}
\end{table}

\subsection{Compositional Generalization and Resilience to Depth}
\label{subsec:compositional}

A critical metric of mechanistic reasoning---as opposed to statistical pattern matching---is \textit{compositional generalization}: the model's ability to maintain performance as the number of required reasoning steps increases. We formalize this via the \textit{normalized degradation rate} $\delta$, defined as the percentage-point accuracy loss per additional hop:
\begin{equation}
    \delta = \frac{Acc_{k=3} - Acc_{k=5}}{5 - 3}
\end{equation}
This metric provides a hop-normalized measure of reasoning fragility that enables comparison across models with different baseline accuracies. Table~\ref{tab:degradation} reports $\delta$ for each evaluated model, computed directly from the accuracy values in Table~\ref{tab:results_table}.

\begin{table}[htbp]
\centering
\caption{Normalized degradation rate $\delta$ (percentage points per hop) across models, computed from Table~\ref{tab:results_table}. Lower values indicate greater compositional robustness.}
\label{tab:degradation}
\begin{tabular}{lcc}
\toprule
\textbf{Model} & \textbf{$Acc_{3} - Acc_{5}$ (pp)} & \textbf{$\delta$ (pp/hop)} \\
\midrule
Gemini 3.1 Pro & 4.7 & 2.35 \\
Qwen 14B (Base) & 4.8 & 2.40 \\
Qwen 14B (SFT) & 3.9 & 1.95 \\
Qwen 14B (SFT + RL) & \textbf{2.7} & \textbf{1.35} \\
\bottomrule
\end{tabular}
\end{table}

\begin{figure}[htbp]
    \centering
    \begin{tikzpicture}
\begin{axis}[
    width=0.92\textwidth,
    height=8.5cm,
    xlabel={\textbf{Reasoning Hop Depth}},
    ylabel={\textbf{Accuracy (\%)}},
    xmin=0.3, xmax=3.7,
    ymin=74, ymax=94,
    xtick={1,2,3},
    xticklabels={\textbf{3-Hop}, \textbf{4-Hop}, \textbf{5-Hop}},
    ytick={74,76,78,80,82,84,86,88,90,92,94},
    yticklabel={\pgfmathprintnumber{\tick}\%},
    legend style={
        at={(0.5,-0.20)},
        anchor=north,
        legend columns=2,
        font=\small,
        draw=gray!30,
        fill=white,
        fill opacity=0.95,
        rounded corners=2pt,
        /tikz/every even column/.append style={column sep=15pt},
    },
    grid=major,
    grid style={line width=0.3pt, draw=gray!20, dashed},
    tick align=outside,
    tick style={color=gray!50},
    axis line style={gray!50, line width=0.6pt},
    label style={font=\small},
    tick label style={font=\small},
    mark size=3.5pt,
    axis on top,
    clip=false,
]


\addplot[
    color=blue!60!black,
    dashed,
    line width=1.4pt,
    mark=*,
    mark options={fill=blue!60!black, solid},
    mark size=3pt,
] coordinates {(1,81.2) (2,76.8) (3,76.4)};
\addlegendentry{Qwen 14B (Base)}

\addplot[
    color=red!70!black,
    solid,
    line width=1.4pt,
    mark=square*,
    mark options={fill=red!70!black},
    mark size=3pt,
] coordinates {(1,88.7) (2,88.3) (3,84.8)};
\addlegendentry{Qwen 14B (SFT)}

\addplot[
    color=violet!80!black,
    solid,
    line width=1.8pt,
    mark=diamond*,
    mark options={fill=violet!80!black},
    mark size=3.8pt,
] coordinates {(1,90.7) (2,89.9) (3,88.0)};
\addlegendentry{Qwen 14B (SFT + RL) --- Ours}

\addplot[
    color=teal!80!black,
    dashdotted,
    line width=1.4pt,
    mark=triangle*,
    mark options={fill=teal!80!black, solid},
    mark size=3.5pt,
] coordinates {(1,86.9) (2,85.2) (3,82.2)};
\addlegendentry{Gemini 3.1 Pro}


\node[below=5pt, font=\scriptsize\bfseries, color=blue!60!black] at (axis cs:1,81.2) {81.2};
\node[below=5pt, font=\scriptsize\bfseries, color=blue!60!black] at (axis cs:2,76.8) {76.8};
\node[below=5pt, font=\scriptsize\bfseries, color=blue!60!black] at (axis cs:3,76.4) {76.4};

\node[left=6pt, font=\scriptsize\bfseries, color=red!70!black] at (axis cs:1,88.7) {88.7};
\node[below =4pt and 3pt, font=\scriptsize\bfseries, color=red!70!black] at (axis cs:2,88.3) {88.3};
\node[right=5pt, font=\scriptsize\bfseries, color=red!70!black] at (axis cs:3,84.8) {84.8};

\node[above=5pt, font=\scriptsize\bfseries, color=violet!80!black] at (axis cs:1,90.7) {90.7};
\node[above=5pt, font=\scriptsize\bfseries, color=violet!80!black] at (axis cs:2,89.9) {89.9};
\node[above=5pt, font=\scriptsize\bfseries, color=violet!80!black] at (axis cs:3,88.0) {88.0};

\node[below=5pt, font=\scriptsize\bfseries, color=teal!80!black] at (axis cs:1,86.9) {86.9};
\node[below=5pt, font=\scriptsize\bfseries, color=teal!80!black] at (axis cs:2,85.2) {85.2};
\node[below=5pt, font=\scriptsize\bfseries, color=teal!80!black] at (axis cs:3,82.2) {82.2};

\draw[<->, thick, gray!60] (axis cs:0.55,81.2) -- node[left, font=\tiny, fill=white, inner sep=1pt] {+9.5} (axis cs:0.55,90.7);
\draw[<->, thick, gray!60] (axis cs:3.45,76.4) -- node[right, font=\tiny, fill=white, inner sep=1pt] {+11.6} (axis cs:3.45,88.0);

\end{axis}
\end{tikzpicture}
    \caption{Accuracy across reasoning hop depths. The Qwen 14B (SFT+RL) model exhibits minimal degradation at higher complexities, significantly outperforming both its baseline and Gemini 3.1 Pro.}
    \label{fig:results_hops}
\end{figure}
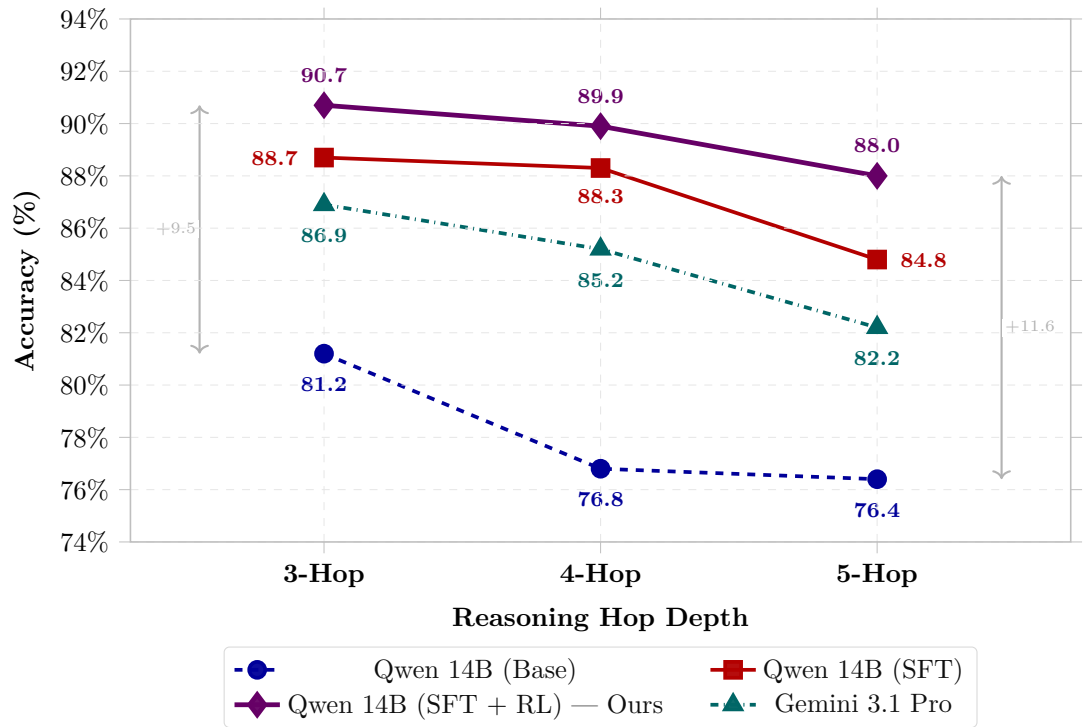

As shown in Fig.~\ref{fig:results_hops} and Table~\ref{tab:degradation}, the SFT+RL model achieves a degradation rate of 1.35~pp/hop, representing a 44\% reduction relative to the base model ($\delta = 2.40$) and a 43\% reduction relative to Gemini 3.1 Pro ($\delta = 2.35$). This reduction is notable because the SFT+RL model was never explicitly trained on 3-, 4-, or 5-hop QA items; its RL phase operated exclusively on 2-hop prompts. The observed resilience to depth, therefore, constitutes \textit{zero-shot compositional transfer}: The model learned to chain axiomatic steps during 2-hop RL and generalized this chaining mechanism to longer, unseen reasoning depths.

The contrast between SFT-only and SFT+RL models isolates the mechanism underlying this resilience. Even though SFT improves absolute accuracy at all depths, it does not fundamentally alter the degradation curve: $\delta_{\text{SFT}} = 1.95$ remains comparable to $\delta_{\text{Base}} = 2.40$. This indicates that supervised learning on multi-hop QA items primarily expands factual coverage without fully internalizing the compositional operator needed to reliably chain facts across extended reasoning paths. The RL phase, which explicitly rewards path-aligned reasoning traces via $R_{\text{path}}$, reduces $\delta$ from 1.95 to 1.35---a 31\% further reduction---indicating that path-derived rewards install a distinct chaining mechanism beyond what SFT provides. This distinction between outcome supervision (SFT) and process supervision (RL with path rewards) is consistent with findings in the alignment literature on the superiority of PRMs for complex reasoning~\cite{lightman2023let}.

In neuroscience, many clinically and scientifically relevant questions involve long causal chains, e.g., tracing how a genetic mutation at the molecular level propagates through cellular, circuit, system, and behavior levels. A model whose accuracy degrades steeply with chain length is unreliable for precisely the questions that require the most expertise. The SFT+RL model's plateau behavior---maintaining 88.0\% accuracy at 5 hops versus 90.7\% at 3 hops---indicates that its compositional reasoning mechanism scales with depth rather than fragmenting under it.

\subsection{SFT as a Prerequisite for Effective RL}
\label{subsec:sft_enables_rl}

A natural question is whether RL alone---without prior SFT---can induce the observed compositional reasoning capabilities. Both the theoretical structure of our reward formulation and prior empirical work indicate that it cannot.

Dedhia et al.~\cite{dedhia2025bottom} address this question directly in their analysis of bottom-up primitives as verifiable rewards. While they identify that KG paths can function as localized verifiers providing dense reward signals during RL, their entire empirical framework relies on SFT over full KG paths to first instill structured reasoning; they do not apply RL as a standalone training method. Their discussion explicitly frames the RL formulation as a \textit{future direction} that builds upon the SFT-grounded model, noting that the KG can be viewed as ``a fully simulatable training environment, where reasoning agents can be optimized not only for end-task correctness but also for intermediate trace fidelity''---but only after the model has acquired the domain's primitives through supervised curriculum tuning.

Our reward formulation encodes this dependency formally. First, $R_{\text{path}}$ is gated on answer correctness [Eq.~(\ref{eq:Rpath})]: Any incorrect completion receives zero path alignment feedback. For the base model, an estimated 15-20\% of 2-hop completions would fall into this zero-reward regime, producing a sparse training signal. Second, the base model has not been aligned to the structured \texttt{<think>}/\texttt{<answer>} format; hence, even among correct completions, the path coverage computation would operate on unstructured text, yielding noisy $R_{\text{path}}$ values. Together, these effects would prevent the stable policy improvement GRPO requires. SFT resolves both issues---raising correctness rates and instilling the structured trace format---ensuring that $R_{\text{path}}$ provides a dense, informative gradient from the first RL training step.

\subsection{Comparison to Frontier Generalist Models}
\label{subsec:frontier}

The SFT+RL model outperforms Gemini 3.1 Pro across all hop depths, with the absolute performance gap computed directly from Table~\ref{tab:results_table}: +3.8 pp at three hops (90.7\% vs.\ 86.9\%), +4.7 pp at four hops (89.9\% vs.\ 85.2\%), and +5.8 pp at five hops (88.0\% vs.\ 82.2\%), averaging +4.8 pp overall. 

The performance gap grows monotonically with hop depth, from +3.8 pp at three hops to +5.8 pp at five hops. This widening gap admits a precise interpretation grounded in error propagation. In autoregressive multi-hop reasoning, each step $t$ in a $k$-step chain introduces an independent error probability $p_t$. Assuming a constant per-step error rate $p$ for simplicity, the probability of a fully correct chain is $(1-p)^k$, which decays geometrically with $k$. Fitting $Acc(k) = Acc(3) \cdot (1 - p)^{k-3}$ to the observed accuracy values yields per-step error estimates of $p_{\text{Gemini}} \approx 0.028$ and $p_{\text{SFT+RL}} \approx 0.015$. The specialist model's per-step error rate is 46\% lower, an advantage that compounds multiplicatively with each additional hop, explaining why the accuracy gap widens at greater depths.

This comparison carries implications for the relationship between parameter scale and curriculum design. Our results demonstrate that for specialized scientific reasoning, the \textit{organization} of training data---specifically, its grounding in verified domain structure---can compensate for substantial differences in parameter count. A 14B model with ${\sim}$20k verified triples and a structured curriculum outperforms a multi-trillion parameter model within the target domain.

Two caveats bound this comparison. First, Gemini 3.1 Pro was evaluated zero-shot without neuroscience-specific adaptation; applying a similar KG-grounded curriculum to a frontier-scale model would likely yield higher performance. Our claim is not that small models are universally superior, but that the bottom-up curriculum is a \textit{sufficient} condition for domain-specific superintelligence at the 14B scale. Second, our evaluation is restricted to multiple-choice questions grounded in a single textbook's KG; generalization to open-ended neuroscience reasoning or to knowledge not covered by the source corpus remains untested.

\subsection{Observed Failure Modes}
\label{subsec:failures}

Despite achieving 89.5\% average accuracy, the SFT+RL model fails on approximately 10\% of evaluation items. Inspection of incorrect reasoning traces reveals recurring failure patterns that, even though not yet systematically quantified, provide actionable directions for improvement.

The most frequent failure mode involves \textbf{path deviation at high-degree nodes}: The model initiates a valid reasoning chain but diverges at an intermediate step where the correct node has many outgoing edges, following a plausible but incorrect relation. For example, a model correctly identifies that glutamate is released by cortical pyramidal neurons but then routes to hippocampal long-term potentiation rather than corticostriatal synaptic transmission. This pattern is expected, given that our path pruning strategy (Section~\ref{sec:methodology}) penalizes but does not eliminate hub nodes, leaving residual ambiguity at high-connectivity junctions.

A second recurring mode is \textbf{entity conflation}: The model confuses semantically similar but functionally distinct entities, particularly at higher hop depths. Common observed conflations include parvalbumin-positive versus somatostatin-positive GABAergic interneurons, ventral versus dorsal striatum, and NMDA versus AMPA receptor subtypes. These conflations involve entity pairs that share ontological categories and frequently co-occur in the KG neighborhood, making them difficult to disambiguate without finer-grained subtype information than our current KG encodes.

A third mode involves \textbf{relation misattribution}: The model identifies the correct entities but assigns an incorrect relation type, such as describing a modulatory relationship as inhibitory. This failure is consistent with the fact that our closed relation vocabulary (Appendix~\ref{app:relations}) contains several semantically overlapping relation types (e.g., \texttt{modulates}, \texttt{regulates}, \texttt{inhibits}) that require fine-grained contextual disambiguation.

These observed patterns point toward two concrete avenues for improvement: (1) enriching the KG with hierarchical subtype information to reduce entity conflation, and (2) developing attention-weighted path rewards that impose heavier penalties for deviations at high-degree branching nodes. A systematic annotation study quantifying the distribution of these error types across hop depths would enable more targeted interventions; we identify this as a priority for future empirical work.

\section{Discussion and Limitations}
\label{sec:discussion}

Our empirical findings substantiate the hypothesis that domain-specific superintelligence can emerge from compact LMs when trained via a bottom-up, neurosymbolic curriculum. Whereas contemporary inference-time scaling techniques effectively elicit deeper reasoning from massive, generalized models \cite{snell2024scaling, muennighoff2025s1}, they inherently rely on the model having internalized the requisite domain axioms during pre-training. Our results demonstrate that top-down pre-training over unstructured web corpora is highly inefficient for acquiring the rigid, axiomatic structures necessary for high-stakes scientific reasoning.

The progression from Qwen3-14B Base (78.1\%) to SFT (87.3\%) to SFT+RL (89.5\%) isolates the contribution of each pipeline stage. The +9.2\% gain from SFT confirms the pedagogical density of the KG-derived curriculum, whereas the additional +2.2\% from RL---concentrated disproportionately at higher hop depths---validates the role of path-derived rewards in promoting genuine compositional generalization rather than surface-level pattern matching.

Critically, the comparison against Gemini 3.1 Pro reveals that our 14B-parameter specialist consistently outperforms a multi-trillion-parameter generalist across all reasoning depths, with the gap widening from +3.8\% at three hops to +5.8\% at five hops. This confirms that explicit grounding in the verified topology of $\mathcal{G}_{exp}$ acts as a robust ``compositional bridge'' \cite{kansal2026knowledge}. The utilization of $R_{\text{path}}$ during the GRPO phase ensures that the model is not merely rewarded for arriving at the correct final token, but is strictly supervised on the validity of its mechanistic reasoning. This implicit reward formulation entirely bypasses the need for costly human-in-the-loop annotations and successfully mitigates the reward-hacking phenomena prevalent in standard outcome-based RLHF \cite{gao2023scaling,shrivastava2025}.

Despite these promising results, our framework exhibits several notable limitations. First, the epistemic boundary of the model is strictly defined by the source corpus, $\mathcal{D}$. While extracting a seed graph from \cite{kandel2013principles} ensures foundational accuracy, the resulting model remains ignorant of recent paradigm shifts in neuroscience that are not yet codified in standard textbooks. Dynamically updating $\mathcal{G}_{exp}$ with real-time literature without introducing ontological contradictions remains an open challenge for continual learning \cite{parisi2019continual}.

Second, our pipeline faces computational bottlenecks during the curriculum generation phase. While traversing a multi-relational graph for $k \in \{1, 2, 3, 4, 5\}$ hops is computationally tractable, the number of valid paths scales exponentially as $k$ increases. Attempting to scale this methodology to highly complex, system-level biological simulations ($k > 10$) would result in a combinatorial explosion, necessitating advanced path-pruning heuristics or neural subgraph routing \cite{kipf2016semi}.

Whereas GraphMERT \cite{belova2025graphmert} excellently distills text-based triples, it ignores critical multi-modal information, such as the rich anatomical diagrams and synaptic circuit schematics that are heavily used in neuroscience education \cite{wang2023knowledge}. Incorporating these visual sources could further strengthen the KG's coverage of spatial and structural relationships.

Finally, our evaluation is confined to a single domain (neuroscience) and a single base architecture (Qwen3-14B). Even though the results are consistent with prior findings on KG-driven curricula in other domains \cite{dedhia2025bottom, kansal2026knowledge}, broader validation across additional scientific fields and model families is necessary to establish the full generality of the approach.

\section{Conclusion and Future Work}
\label{sec:conclusion}

In this work, we introduced an end-to-end neurosymbolic pipeline designed to instill expert-level, multi-hop reasoning capabilities into compact LMs. By abandoning the traditional top-down pre-training paradigm in favor of a bottom-up, KG-grounded curriculum, we demonstrated the emergence of domain-specific superintelligence in the field of neuroscience.

Our methodology leveraged GraphMERT to scalably extract and validate a highly reliable KG comprising 9,187 nodes and 19,755 triples from a single authoritative textbook. By systematically traversing this graph, we generated a stratified QA curriculum spanning 1-hop to 5-hop reasoning depths. Using a dual-stage training approach---initial SFT to establish atomic axioms, followed by path-rewarded RL using GRPO---we forced the model to explicitly verify its intermediate reasoning steps against the KG topology. Consequently, our 14-billion-parameter Qwen3 model achieved 89.5\% average accuracy across 3-hop to 5-hop reasoning tasks, outperforming the frontier Gemini 3.1 Pro model by 4.7\% on average and exhibiting markedly flatter performance degradation as reasoning depth increased.

Future work will focus on three directions. First, we aim to expand the modality of the extraction pipeline by integrating vision-language models into the extraction phase, augmenting the text-derived KG with spatial and structural relations extracted directly from textbook diagrams. Second, applying this framework across intersecting scientific disciplines---such as integrating a neuroscience KG with a pharmacological KG---could pave the way for autonomous systems capable of discovering novel neuro-therapeutics through verifiable, multi-hop deductive reasoning. Third, we plan to evaluate the generality of the pipeline across additional model families and parameter scales to characterize the minimum model capacity required for compositional generalization under KG-grounded supervision.

\vspace*{1mm}
\noindent
\textbf{Acknowledgment:} The authors would like to thank Yuval Kansal, Margarita Belova, and Jiaxin Xiao for their help with using GraphMERT and domain-specific superintelligence tools, and Yuval Kansal for help with preparing the webpage for this project.

\appendix

\section{Knowledge Graph Extraction Prompt}
\label{app:prompt}

The system prompt presented next is used verbatim for all text unit extraction calls during Phase~1. The placeholder \texttt{\{relation\_list\}} is populated at runtime with the JSON-serialized closed-vocabulary relation set; delimiter tokens \texttt{\{tuple\_delimiter\}}, \texttt{\{record\_delimiter\}}, and \texttt{\{completion\_delimiter\}} are instantiated as \texttt{<|>}, \texttt{\#\#}, and \texttt{<|COMPLETE|>}, respectively.

\begin{quote}
\ttfamily\small\setstretch{1.05}
\textbf{-Role-}\\
You are an AI assistant specialized in extracting structured information from neuroscience textbook content to build a knowledge graph about the nervous system, brain function, and neural mechanisms.

\medskip
\textbf{-Goal-}\\
Given neuroscience textbook content, a predefined list of entity types, and a predefined list of relations, identify every entity of those types and the scientifically meaningful relationships explicitly described among them within the text. Extract only information directly stated in the text---do not infer, generalize, or use external scientific knowledge.

\medskip
\textbf{-Output Format-}\\
For each entity:\\
\hspace*{1em}(\textquotedbl entity\textquotedbl\{tuple\_delimiter\}name\{tuple\_delimiter\}type\{tuple\_delimiter\}description)
\{record\_delimiter\}

\smallskip
For each relationship:\\
\hspace*{1em}(\textquotedbl relationship\textquotedbl\{tuple\_delimiter\}source\{tuple\_delimiter\}target\{tuple\_delimiter\}
relation\{tuple\_delimiter\}strength)\{record\_delimiter\}

\noindent where strength: 7 = central, 5 = supporting, 3 = brief mention. Output ONLY tuples. End with \{completion\_delimiter\}.
\end{quote}

\section{Closed-Vocabulary Relation Types}
\label{app:relations}

Table~\ref{tab:relations} enumerates the 40 directed relation types comprising the closed extraction vocabulary used in Phase~1, organized by functional category for interpretability.

\begin{table}[h]
\centering
\caption{Closed-vocabulary relation types used for seed KG extraction.}
\label{tab:relations}
\small
\begin{tabular}{ll}
\toprule
\multicolumn{2}{l}{\textit{Anatomical \& Connectivity}} \\
\midrule
\texttt{part\_of}               & \texttt{contains} \\
\texttt{located\_in}            & \texttt{connected\_to} \\
\texttt{projects\_to}           & \texttt{receives\_input\_from} \\
\texttt{receives\_modulatory\_input\_from} & \texttt{innervates} \\
\texttt{originates\_from}       & \texttt{terminates\_in} \\
\midrule
\multicolumn{2}{l}{\textit{Molecular \& Cellular}} \\
\midrule
\texttt{expressed\_in}          & \texttt{synthesized\_in} \\
\texttt{releases}               & \texttt{binds\_to} \\
\texttt{activates}              & \texttt{inhibits} \\
\texttt{modulates}              & \texttt{regulates} \\
\texttt{transports}             & \texttt{forms\_complex\_with} \\
\midrule
\multicolumn{2}{l}{\textit{Functional \& Representational}} \\
\midrule
\texttt{responds\_to}           & \texttt{fires\_in\_response\_to} \\
\texttt{tuned\_to}              & \texttt{selective\_for} \\
\texttt{encodes\_representation\_of} & \texttt{participates\_in} \\
\texttt{required\_for}          & \texttt{sufficient\_for} \\
\texttt{oscillates\_at}         & \texttt{mediates\_signal\_for} \\
\midrule
\multicolumn{2}{l}{\textit{Causal \& Clinical}} \\
\midrule
\texttt{associated\_with}       & \texttt{causes} \\
\texttt{results\_in}            & \texttt{impaired\_in} \\
\texttt{degenerates\_in}        & \texttt{risk\_factor\_for} \\
\texttt{biomarker\_of}          & \texttt{symptom\_of} \\
\texttt{treated\_by}            & \texttt{diagnosed\_by} \\
\bottomrule
\end{tabular}
\end{table}

\end{document}